\documentclass{article}

\usepackage{arxiv}

\usepackage[utf8]{inputenc} % allow utf-8 input
\usepackage[T1]{fontenc}    % use 8-bit T1 fonts
\usepackage{hyperref}       % hyperlinks
\usepackage{url}            % simple URL typesetting
\usepackage{booktabs}       % professional-quality tables
\usepackage{amsfonts}       % blackboard math symbols
\usepackage{nicefrac}       % compact symbols for 1/2, etc.
\usepackage{microtype}      % microtypography
\usepackage{cleveref}       % smart cross-referencing
\usepackage{lipsum}         % Can be removed after putting your text content
\usepackage{graphicx}
\usepackage{natbib}
\usepackage{doi}

\usepackage[english]{babel} % for Arabic
\usepackage[LAE,T1]{fontenc} % for Arabic

\title{LANS: Large-scale Arabic News Summarization Corpus}

\author{
Abdulaziz Alhamadani{$^\dag$}, Xuchao Zhang{$^+$}, Jianfeng He{$^\dag$}\thanks{~~Corresponding author.}, 
 Chang-Tien Lu{$^\dag$}\\ 
 {$^\dag$}Sanghani Center for Artificial Intelligence and Data Analytics, Virginia Tech, VA, USA\\
   {$^+$} NEC Labs America, NJ, USA\\
 {$^\dag$}\{hamdani, jianfenghe, ctlu\}@vt.edu,
 {$^+$}xuczhang@nec-labs.com
 }

\renewcommand{\shorttitle}{LANS: Large-scale Arabic News Summarization Corpus}

\hypersetup{
pdftitle={A template for the arxiv style},
pdfsubject={q-bio.NC, q-bio.QM},
pdfauthor={David S.~Hippocampus, Elias D.~Striatum},
pdfkeywords={Arabic Language, dataset, Text Summarization},
}
\begin{document}
\maketitle

\begin{abstract}
	%\lipsum[1]
	Text summarization has been intensively studied in many languages, and some languages have reached advanced stages. Yet, Arabic Text Summarization (ATS) is still in its developing stages. Existing ATS datasets are either small or lack diversity. We build, LANS, a large-scale and diverse dataset for Arabic Text Summarization task. LANS offers 8.4 million articles and their summaries extracted from newspapers websites’ metadata between 1999 and 2019. The high-quality and diverse summaries are written by journalists from 22 major Arab newspapers, and include an eclectic mix of at least more than 7 topics from each source. We conduct an intrinsic evaluation on LANS by both automatic and human evaluations. Human evaluation of 1000 random samples reports 95.4\% accuracy for our collected summaries, and automatic evaluation quantifies the diversity and abstractness of the summaries. The dataset is publicly available upon request.
\end{abstract}

% keywords can be removed

\section{Introduction}
Every day there is an abundant amount of text published on the internet, such as news articles, scientific papers, product reviews, and blogs. Therefore, the need for text summarization is compelling to make use of this information overload. For a summarized text, a good one should be concise and include the main information of the original text \citep{radev2002introduction}. For some languages like English, the field has developed rapidly and achieved competitive results\citep{zhang2020pegasus, lewis2019bart,dou2020gsum}. Unlike English, the field in Arabic has been slowly and fairly developing in the past few years; thus, it has not reached its advanced shape. In the field of Arabic Text Summarization (ATS)\citep{belkebir2015supervised,al2015lexical,fejer2014automatic,abu2020arabic,el2021automatic}, the dearth of a diverse and large summarization dataset is one of the main existing difficulties that ATS researchers encounter\citep{al2016automatic,elsaid2022comprehensive}.

\begin{figure}
\centering
\includegraphics[width=0.5\textwidth]{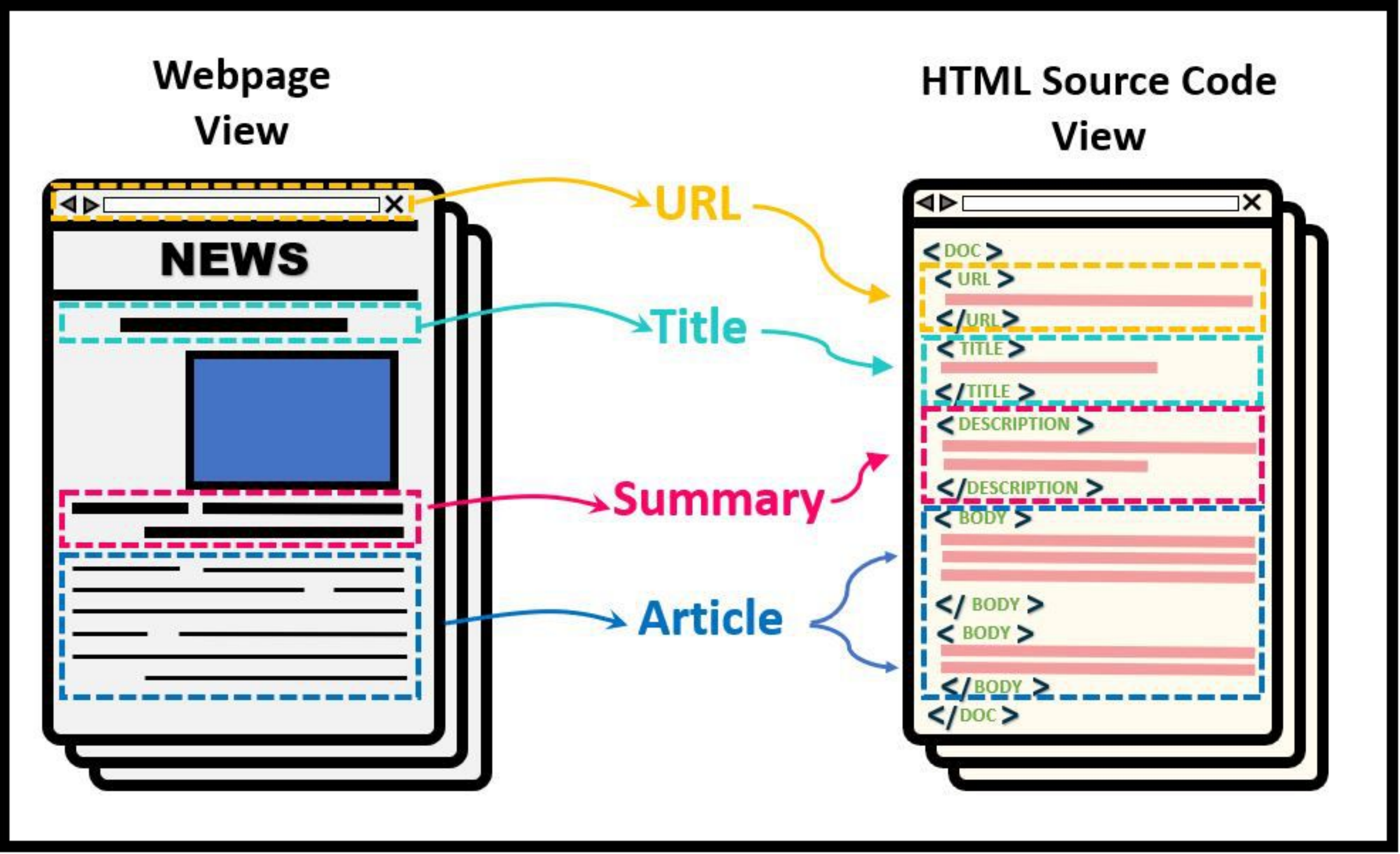}
\caption{The Webpage view (left) shows a typical news article view. The summaries are extracted from the HTML source code view's (right) metadata (\textit{og:description}).  } \label{figure1}
\end{figure}

Concerted efforts have been made to overcome those challenges by building various Arabic datasets for the task such that EASC\citep{el2010using}, Kalimat\citep{el2013kalimat}, TAC2011\citep{el2014multi}, ANT\citep{chouigui2021arabic}, and XL-Sum\citep{hasan2021xl}, but those datasets have limitations in terms of diversity or size. Therefore, the demand for a diverse and large-scale dataset is crucial to advance the ATS field. The diversity in the ATS dataset is in twofold. The first kind of diversity exists in the Modern Standard Arabic (MSA). Even though 22 countries use MSA as an official standard language, each country has its own dialects (Dialectal Arabic) for communication. Each country's dialects have some effects on the MSA style of writing and the choice of words. For example, in a sentence describing the rounds of a soccer match, Morrocan MSA would use the word "\bgroup\beginR\fontencoding{LAE}\selectfont أطوار\endR\egroup"  for "rounds" and "\bgroup\beginR\fontencoding{LAE}\selectfont المواجهة\endR\egroup" for word "the match" while Saudi MSA would use "\bgroup\beginR\fontencoding{LAE}\selectfont أشواط\endR\egroup" and "\bgroup\beginR\fontencoding{LAE}\selectfont المباراة\endR\egroup" respectively. Second, there is diversity in news categories. Each newspaper has different news topics, such as finance, politics, sports, health, local, international news, and more. Not all ATS datasets include both diversity aspects in one dataset (see Table~\ref{compare-table}). Thus, it is essential to build a dataset that considers both types of diversity.

In terms of size, the available ATS datasets contain a range of 100 to 41,000 training samples, which make them too small to fully train a summarization model. The performance in summarization models evidently relies on a substantial amount of applicable training samples\citep{volske-etal-2017-tl,grusky2018newsroom,zhang2020pegasus, lewis2019bart,dou2020gsum}. Thus, we expect a large-scale dataset that is provided in this work.

To overcome the current limitations in diversity and size, we introduce a new ATS dataset (\textbf{LANS}) that includes both types of diversity and large-scale to present new opportunities to ATS models and improve their summary accuracies. To achieve MSA diversity, that is the variety of each Arab country dialects' impact on its MSA, LANS encompasses 19  Arab countries and collected articles along with their summaries of 22 popular newspapers (see Table~\ref{stats}). For the diversity of text categories, we consider all available news categories of each source in our ATS dataset. Thus, LANS ensures both types of diversity of MSA among the Arab countries. To overcome the size limitation, LANS provides more than 8 million news articles along with their summaries. LANS's substantial amount of articles and their summaries, plus the diversity in MSA sources and categories make it a worthy resource for ATS models. 

LANS exploited the metadata of newspapers’ archives to extract and build the dataset. In Figure.~\ref{figure1}, a high-level example is shown to demonstrate where the collected information originated from two parallel views: the webpage view and its HTML source code view. The webpage view shows what a reader sees when reading a news article: the URL, title, bold part or the abstract sentence/s, and article bodies. LANS pursues the metadata attributes from the HTML source code - specifically (\textit{og:description}) to extract the summaries from. In the webpage view, the summaries lie either in bold text or before the article's paragraphs. In the HTML source code view, the summaries lie in the metadata attributes, in our case between (\textit{og:description}) tags, which we extracted as the news articles' summaries. After the extraction, we cleaned and filtered 11M news articles to present 8.4M articles along with their summaries. 

To quantify the quality of the collected summaries and examine their summarization properties, we conducted an automatic evaluation based on 3 common metrics. Moreover, we corroborated the evaluation with the human evaluation of 1000 samples to verify the accuracy of using the abstract from the HTML source code's metadata as a summary. The human evaluation verifies that using the summary available in the metadata has a 95.4\% accuracy. Considering the large size of LANS, 8.4 million, LANS can benefit the ATS field, because large datasets improve NLP tasks, such as numerous training samples for pre-trained models\citep{zhang2020pegasus, lewis2019bart}. Besides, both types of diversities create opportunities for researchers to construct more accurate ATS models.  

\begin{table*}[htb]
\scriptsize
%\small
\begin{center}
  \centering
\begin{tabular}{cccccccccc}
\hline
ID & Newspaper    & Country    & From & \multicolumn{1}{c|}{Articles} & ID & Newspaper      & Country      & From                   & Articles  \\ 
\hline\hline
1  & Elkhabar     & Algeria    & 2014 & \multicolumn{1}{c|}{78201}    & 12 & Hespress       & Morroco      & 2007                   & 91357     \\
2  & Alwasat      & Bahrain    & 2013 & \multicolumn{1}{c|}{23860}    & 13 & Alwatan        & Oman         & 2014                   & 130067    \\
3  & Gate Ahram   & Egypt      & 2016 & \multicolumn{1}{c|}{315655}   & 14 & Alquds         & Palestine    & 2015                   & 88313     \\
4  & Youm7        & Egypt      & 2008 & \multicolumn{1}{c|}{2039818}  & 15 & Alquds-UK      & Palestine    & 2013                   & 349439    \\
5  & Albayan      & Emirates   & 1999 & \multicolumn{1}{c|}{1137188}  & 16 & Alwatan        & Qatar        & 2016                   & 214405    \\
6  & Almadapaper  & Iraq       & 2009 & \multicolumn{1}{c|}{105925}   & 17 & Aljazira       & Saudi Arabia & 2001                   & 809445    \\
7  & Aldustoor    & Jordan     & 2000 & \multicolumn{1}{c|}{601372}   & 18 & Alryiadh       & Saudi Arabia & 2004                   & 1004893   \\
8  & Annahar      & Kuwait     & 2007 & \multicolumn{1}{c|}{575482}   & 19 & Alsudan Alyoom & Sudan        & 2016                   & 104439    \\
9  & Alakhbar      & Lebanon   & 2006 & \multicolumn{1}{c|}{222215}   & 20 & Zamanalwsl     & Syria        & 2007                   & 128785    \\
10 & WAL          & Libya      & 2013 & \multicolumn{1}{c|}{141898}   & 21 & Alssabah       & Tunisia      & 2011                   & 166137    \\
11 & Sahara Media & Mauritania & 2009 & \multicolumn{1}{c|}{11982}    & 22 & Almasdar       & Yemen        & 2009                   & 102608    \\ \hline
   &              &            &      &                               &    &                &              &Total  & 8,443,484 \\ \hline
%   \bottomrule
\end{tabular}
\caption{Overall statistics of the collected articles}
\label{stats}
\end{center}
\end{table*}

Our main contributions are as follows: (1) We curate LANS, a large-scale ATS dataset of 8.4 million Arabic news articles paired with their summaries written by journalists between 1999 to 2019. To our knowledge, it is the largest to date. (2) LANS is collected from 22 reputable Arab newspapers to achieve high quality of diversity in MSA, and for each source, there are at least 7 topics to achieve diversity in categories. 
(3) To quantify the intrinsic quality of LANS, a human evaluation is conducted on 1000 random samples and verifies 95.4\% accuracy of the summaries. Plus, the automatic evaluation on the whole dataset quantifies the abstractness and properties of the summaries.

%\lipsum[2]
%\lipsum[3]

\section{Related Work (Existing Datasets)}
\label{rel_work}
To the best of our knowledge, Lakhas~\citep{douzidia2004lakhas} is considered one of the early works to build an ATS model. Due to the lack of ATS datasets at that time, Douzidia et al. translated (DUC)\footnote{An English text summarization dataset of news paired with human summaries. \url{https://duc.nist.gov/}} dataset, from English to Arabic for their ATS model's evaluation~\citep{douzidia2004lakhas}. The translation used machine translation at that time which was not as accurate and advanced as these days, and that had a negative impact on the results. Moreover, other ATS models built their own datasets to evaluate their models\citep{al2020arabic}. Consequently, researchers built Arabic ground-truth summaries over the past years, and this section mentions the major ones.

\textbf{The Essex Arabic Summaries Corpus (EASC) Dataset.}
EASC \citep{el2010using} is an ATS dataset, where each summary is  extracted from the texts by Mechanical Turk. Its text source is two Arabic newspapers (Alrai and Alwatan) and the Arabic language version of Wikipedia. As a result, it contains 153 Arabic articles and 765 summaries (5 summaries per article). In short, EASC has high-quality human-generated summaries but it is too small and lacks diversity.  %But it is confusing for some researchers to evaluate their results with other system because there are 5 candidate summaries for each article. 

\textbf{Kalimat Dataset.} %Because collecting data in EASC was costly,
El-Haj et al. worked on a dataset called Kalimat\citep{el2013kalimat}. It has 20,291 extractive Single-document and multi-document system summaries, and includes only 6 categories. It has been collected from only one source, which is Alwatan newspaper from Oman. The single-document summaries are generated based on their model Gen-Summ which inputs the article and its first sentence, then outputs the extractive summary. The multi-document summaries were generated for each 10, 100, and 500 articles in different categories. The generated summaries also lack human evaluation of the summaries.

\textbf{Arabic News Texts Corpus (ANT) and XL-Sum.} ANT ~\citep{chouigui2021arabic}, and XL-Sum\citep{hasan2021xl} are the most recent works. ANT collected 31,798 documents paired with summaries using RSS feeds from 5 Arab news sources: AlArabiya, BBC, CNN, France24, and SkyNews, while XL-Sum collected 40,327 only from BBC. ANT includes 6 categories%: culture, economy, international news, Middle East news, sports, and technology
, while XL-Sum reported none. Unlike ANT, LANS utilized the HTML source code \textit{og:description} tag to collect the summaries which is similar to \citep{grusky2018newsroom}. ANT is evaluated on several extractive summarization methods such as LexRank, TextRank, Luhn and LSA. XL-Sum fine-tuned mT5 on their dataset and randomly sampled 500/500 development and test set respectively. Besides, they conducted human evaluation on 250 random samples. When compared to our LANS, our work collected nearly 8 million articles with summaries from 19 Arab countries local newspapers. Moreover, experts evaluated 1000 random summaries from LANS to substantiate the validity of the summaries. 
%I don't know if I should add XL-SUM or not

\begin{table*}[htb!]
\small
\begin{center}
%\begin{tabular}{|p{1cm}|p{1cm}|p{1cm}|p{1cm}||p{1cm}|}
\begin{tabular}{ccccc}

       \hline
       Corpus   & \# of documents & MSA Diversity & Category Diversity   & Human Evaluation \\
       \hline\hline
		EASC     & 153             & $\times$                & $\times$         & $\checkmark$    \\
		KALIMAT  & 20291           & $\times$            & $\checkmark$         & $\times$        \\
		ANT     & 31798           & $\times$            & $\checkmark$        & $\times$        \\
		XL-Sum &   40327  & $\times$            & $\checkmark$     & 250  \\
		LANS & $>$ 8 millions  & $\checkmark$            & $\checkmark$     & 1000  \\
       \hline

 \end{tabular}
 \caption{Arabic Text Summarization Datasets comparison}
 \label{compare-table}
 \end{center}
 \end{table*}

\section{LANS Dataset}
\label{LANS_dataset}
Collecting, processing, extracting, processing, and building any dataset are meticulous work. This section details how LANS is collected starting from the scraping process to building the dataset and how it is shaped for public use.

\subsection{Data Collection}
\label{collection}
Our main goal is to improve the ATS field by collecting and building the largest and most diverse ATS dataset. 
% Before collecting the articles, all 22 Arab countries were planned to be searched for all their available newspapers on the web.
We collect newspapers from 19 countries~\footnote{There are 22 Arab countries, but 3 of them: Djibouti, the Comoros Islands, and Somalia, lack Arabic data and reliable newspapers}.
For consistency and fairness of data collection, all the TV news channels' websites are excluded, like Alarabiya, Aljazeera, Arabic CNN, and Arabic BBC because they are primarily established as TV news channels. 
To make our data sources comprehensive and trustworthy, we collected and listed approximately all the  reliable newspapers for each country.
% For each country, a comprehensive research is conducted to list all the possible reliable newspapers for all the countries. 
For instance, we listed 18 reputable newspapers in Saudi Arabia. After analyzing the newspapers, we then ranked them by assigning the highest priority to the newspaper with the longest publishing history.
% and increase the reliability of choosing a newspaper, the newspapers with good history of publishing were given higher priority over the modern ones even though they have more views from search engines. 

Next, we only select the newspapers if their content passes certain criteria:  \textbf{a-} history of published articles (archive),  \textbf{b-} diversity in categories, and  \textbf{c-} availability of the newspaper's summary in the metadata. \textbf{History of published articles (archives):} each newspaper's website is inspected to examine if it has a considerable historical electronic archive to reestablish the long-history versions of a newspaper. An old reputable newspaper can be given a lower rank over a modern one if the latter has a longer historical e-archive. Thus, LANS has collected data from 1999 to 2019 see Table~\ref{stats}.  \textbf{Diversity in categories: } 
% in addition to the archive, the variety of categories was one of the criteria of choosing a newspaper. 
% another major criteria is that 
a newspaper should contain a variety of topics or categories (at least 7), for example, local news, international news, politics, economy, religion, culture, health, sports, art, technology, and so on. \textbf{Availability of the summary in the metadata: } the metadata of a document has the hidden information of an article. The summary of an article written by the author initially lies in the metadata and also can appear in bold on the webpage or ahead of the article. The availability of the summary published by the author/journalist is the major factor in selecting the newspaper. Only the newspapers with provided summaries in the metadata are selected. 

The aforementioned criteria narrow down the list of the reliable newspapers, shown in Table~\ref{stats}. As a result, 22 popular newspapers of 19 Arab countries have been selected for the next step from the period of time between 1999 to 2019. The wide variety of the data sources can significantly benefit the diversity of the summaries.

\subsubsection{Data Scraping}
\label{scraping}
Since there are 22 newspaper websites to be scraped, it is necessary to customize a code for each of them. Each code identifies the patterns, the selectors, and the URLs to be scraped. The main information scraped from each news article are the following: URL, title or (headline), article, and finally the summary or (the metadata from \textit{og:description}). An example is shown in Table~\ref{scrape_table}, which shows the scraped information from an article's webpage. For reproducibility, \textit{Scrapy} was ideal, in our case scenario, for implementing recurring and large-scale web scraping projects. Besides, \textit{Scrapy} supports different built-in data output such as JSON, XML, and CSV.

\begin{table}[]
\scriptsize
\caption{Scraped information from an Article}
  \label{scrape_table}
  \centering
\begin{tabular}{p{1.5cm}|p{11.5cm}}
\toprule
Selector & Scraped info                                                      \\
\midrule
URL      & \url{http://www.alwasatnews.com/news/1196668.html} \\
Title    &  \bgroup\beginR\fontencoding{LAE}\selectfontبالصور... المرخ الخيرية تنظم حملة تنظيف لمقبرة القرية\endR\egroup                                           \\
Article  & \bgroup\beginR\fontencoding{LAE}\selectfont قام المشاركون بإزالة الأشجار والأوساخ الضارة وتقليم الأشجار، وقد شهدت الحملة مشاركة من الأهالي من جميع الفئات العمرية، بالإضافة لأعضاء مجلس إدارة الجمعية.
من جانبه، قال رئيس لجنة شئون القرية والمقبرة في الجمعية مصطفى عبدالنبي إن الحملة تأتي استكمالاً لعملية التطوير شامل للمقبرة، حيث تستعد اللجنة للبدء بالمرحلة السادسة من عملية تطوير المقبرة والتي ستشمل عمل كراسي للمظلة ورصف الطريق المؤدي من المغتسل إلى المظلة ونقل خزان الماء الرئيسي من موقعه الحالي إلى الجهة الشرقية للمغتسل وإصلاح واستكمال شراء الاحتياجات، بالإضافة إلى متابعة الخطة التطويرية بالتنسيق مع إدارة الأوقاف الجعفرية، هذا وأثنى على نشاط المشتركين في الحملة، كما قدم شكره لجميع أبناء القرية لتعاونهم لإنجاح حملة تنظيف المقبرة. endR\egroup
                                        \\
\hline
Summary  & \bgroup\beginR\fontencoding{LAE}\selectfontنظمت لجنة شئون القرية في جمعية المرخ الخيرية الاجتماعية، تزامناً مع رأس السنة الميلادية، حملة تنظيف لمقبرة القرية تحت شعار استثمر وقتك لآخرتك، صباح أمس الأحد1  يناير كانون الثاني 7102\endR\egroup \\
\bottomrule
\end{tabular}
\end{table}

\subsection{Building LANS Dataset}
For the collected data to be curated so it preserves a good quality for reuse and evaluation, we detail how the data is extracted, cleaned, and preprocessed. 

\subsubsection{Data Extraction}
\label{extraction}
 Among the data formats for retrieval, the most convenient format to preserve data quality is XML. The extracted data is stored in a tree structure. Each newspaper has a dataset formatted as the following: "Item" is the root node of the tree. The root has many child nodes "Items". Each "Items", a child node, holds the extracted data of a single document (a newspaper article). The child node, "Items", has 4 child nodes of its own named: Address, Title, Article, and Summary. Each child node of the parent "Items" (Address, Title, Article, and Summary) has 1 or more grandchild nodes depending on the actual values extracted from an article's webpage. 
The data in this stage is not considered clean nor reliable because it contains many errors that could impact the quality of LANS. Errors can be extraneous or foreign characters, empty values, HTML code, or other common text errors. Thus, we need to clean the data. Plus, to better utilize the data in the XML files, we need to preprocess the data for the evaluation process.
% In this stage, a close examination is needed and so is pre-processing for the extracted data. In order to pre-process the data and provide detailed analysis of the data, the data needs to be extracted from the XML files.

\textbf{Data cleaning: }Initially, more than 11 million articles and their metadata are scraped. The data is laboriously examined to ensure whether the extracted articles are error-free content or not, and to ensure their validity for usage. One of the main errors was the collected articles with missing content. There are some reasons for that. One of the reasons is that many articles contain only images or videos without any textual content, because they are types of news that only report pictures or videos. The other reason for missing content is mistakes from the HTML pages, or content stored under a different selector. All articles with the mentioned errors are removed. Moreover, to clean the other errors the normalization step in the preprocessing steps below is performed. In short, the removed articles may have no title, article, or valid data. After removing all the unusable articles, the number has dropped from 11,115,932 to 8,443,484 articles. After this step, the data is stored in its final XML tree format. %The XML files of some newspapers are very large to be opened or processed. The sizes of the large XML files range from 3 GB to 7 GB. Opening or processing such files is difficult but not implausible. Nevertheless, the large XML files are chunked to facilitate the work for researchers.

\subsubsection{Preprocessing}
\label{preprocess}
Even though the data is clean at this stage, it requires preprocessing for ATS evaluation process, due to the complex and rich nature of Arabic language. 
% In our case the preprocessing is necessary for the automatic evaluation process.
% Due to the complex and rich nature of Arabic language, preparing Arabic text for ATS models requires several steps.  
The steps involve normalization, segmentation, removal of stop words, and lemmatization; in that order. This stage in Arabic is the primary stage to prepare the text for processing and transform the input text into a unified representation.

The normalization step %, as mentioned earlier, has a major effect on the quality of data, since it further 
cleans the data and removes many extraneous texts. It removes extra white spaces or tabs, foreign irrelevant characters, non-letters, and diacritics. It also replaces certain Arabic characters with a certain single character to normalize the differences in characters. Normalization also removes the "Tatweel" (character stretching)  \citep{ayedh2016effect}. For tatweel, a word that appears in this format   "\bgroup\beginR\fontencoding{LAE}\selectfontتــمـــديـــد\endR\egroup" is going to be replaced with "\bgroup\beginR\fontencoding{LAE}\selectfontتمديد\endR\egroup"

Segmentation or tokenization are commonly used interchangeably. The segmentation process is applied to segment the article into sentences and prepare for the next steps. We use the Natural Language Toolkit(NLTK) \citep{Loper02nltk:the} to tokenize sentences and words. We are aware that some scholars weigh tokenization differently such as when tokenization breaks the words into constituent prefix(es), stem, and suffix(s)\citep{mubarak2017build,abdelali2016farasa,el2015enhancing,pasha2014madamira}. However, ATS lemmatization accomplishes the intended purpose of the other definition of Arabic tokenization.

Stop words have a major impact on text summarization because they impact the length of the articles and summaries, and increase the frequency of words which in both cases would change the weights of sentences\citep{el2017effects,al2014extractive}. To remove the stop words, we used a list of stop words prepared by Abu El-khair et al\citep{el2017effects} which contains 1377 words. 
 
For our evaluation, the final and most crucial step for preprocessing the text is Lemmatization. This step can improve the accuracy of the summarization and evaluation process. Lemmatization is the process of reducing words to their basic root by removing the attached affixes of words.  LANS dataset does not store the data in the lemmatized format, because lemmatization is usually used in the training or testing on the original data. Many lemmatizers are considered such as Alkhalil\citep{boudchiche2019hybrid}, ISRI (Khoja)\citep{el2015enhancing}, Madamira \citep{pasha2014madamira}, CAMeL\citep{obeid2020camel}, but only Farasa\citep{mubarak2017build,abdelali2016farasa} is applied because it outperforms the state-of-the-art CAMel by a slight margin and its fast performance on large-scale datasets. Following all the mentioned steps, the dataset is passed for automatic evaluation (see sec~\ref{evaluation}).

\section{LANS Description}
\label{others}

LANS builds 8,443,484 articles and their summaries from 22 newspapers of 19 Arab countries dated from 1999 to 2019. The high-level overall statistics in Table~\ref{stats} show that some newspapers have more data than the others. This does not undermine any country's newspapers. Among the newspapers with a long history of journalism, most of them have been published on physical newspapers before newspapers become digitalized. The dates of collection reflect how much data is available in the e-archive for each newspaper. For instance, Gate Ahram newspaper from Egypt \citep{Gate_Ahram} is established in 1875 and has been published since then. However, the available e-archive for the newspaper starts from 2016. Each newspaper's webpage has its own e-archive and its own progress over time. This is why the variations of collection dates exist.  

LANS encompasses 19 Arab countries for MSA diversity.
% This research focuses on several crucial aspects in the field of ATS. 
One of the overlooked aspects of diversity in Arabic is the diversity of MSA in the Arab countries. It is true that all the newspapers in the Arab countries use the same MSA, but events, culture, and use of vocabulary are different from one country to another. Therefore, it is necessary to collect such diverse data from each country. To achieve MSA diversity in LANS, our dataset encompasses 19 Arab countries - except for the Comoros Islands, Djibouti, and Somalia because of the scarcity of data in their newspapers.  

Further, LANS provides a wide-ranging topic variety. The collected data from each country covers different categories, and some newspapers have more categories than others, which enhances the diversity of categories in LANS. Some newspapers have only a few categories (not less than 7), while some others have more than 9 categories including local news, international, political, financial, society, sports, technology, art, health, and religious news articles. This category diversity is one of the features of LANS. It allows researchers to not only create subdatasets, but also create sub-subdataset of any of the subdatasets. For example, a subset can be all articles/summaries from Saudi Arabia. Then, a sub-subdataset can be the local news categories from the subset of Saudi Arabia articles/summaries. This type of diversity can be created from LANS. 

The dataset is chunked into separate XML files, each file is under 2 GB to make it easier to load and process. The total size of the whole dataset is 32GB. Each country's dataset is a subset of the whole dataset, and researchers have the freedom to choose a subset or several subsets (by specific countries) to train and evaluate ATS models.

\begin{table}[ht]
\scriptsize
\caption{Table presents a sample of two summaries from LANS and mT5-based pipeline.}
  \label{sample_com}
  \centering
\begin{tabular}{p{1.5cm}|p{11.5cm}}
\toprule
 \multicolumn{1}{c}{}     &   \multicolumn{1}{c}{Summary}    \\
\midrule
\multicolumn{1}{c}{LANS}      & 
\bgroup\beginR\fontencoding{LAE}\selectfont من المقرر الكشف عن اسماء اهم خمسة مرشحين لجائزة افضل لاعب كرة قدم في افريقيا للعام الحالي غداً الأحد ويتوقع ان يكون كابتن منتخب نيجيريا جاي جاي اوكوتشا من بين اقوى المرشحين للجائزة السنوية.
\endR\egroup
 \\
 \midrule
\multicolumn{1}{c}{mT5-based pipeline}  & 
\bgroup\beginR\fontencoding{LAE}\selectfont من المقرر الكشف عن أفضل خمسة مرشحين لأفضل لاعب كرة قدم في أفريقيا لهذا العام هذا الأحد، ومن المتوقع أن يكون الكابتن المنتخب لنيجيريا جاي أوكوكوشا من بين أقوى المرشحين للجائزة السنوية. أوكوشا، المرشح الرئيسي لأفضل لاعب أفريقي.
\endR\egroup
\\
\bottomrule
\end{tabular}
\end{table}

\section{Experiment}
\label{experiment}
Since the ATS field is still under-researched for \textit{abstractive} summarization, it is difficult to achieve multiple comparisons among the available works. Therefore, we created a translate-summarize-translate pipeline from the available pretrained state-of-the-art multi-language models such mT5\citep{xue2020mt5}, mBART\citep{tang2020multilingual}, and CRISS\citep{tran2020cross}. For our experiment, we chose mT5 because of its wide coverage of 101 languages and support for 41 languages. The model is utilized to generate summaries of the 1000 randomly sampled articles, and then compare them with LANS ground-truth summaries using ROUGE-N.  In a high-level description, the pipeline inputs the preprocessed samples as mentioned earlier in section~\ref{preprocess}, translates the articles (Arabic $\rightarrow$ English), generates summaries from the translated articles, then translates the generated summaries (English $\rightarrow$ Arabic) for evaluation. The model for each step of the pipeline will be given later.

Some of the pipeline steps to generate automatic text summaries are tuned to adapt Arabic language. Firstly, we preprocess the text, as detailed in section~\ref{preprocess}. Secondly, we translate the articles from Arabic to English. We apply OPUS-MT \citep{tiedemann-thottingal-2020-opus} project. OPUS-MT is based on Marian-NMT \citep{mariannmt}, a state-of-the-art transformer-based Neural Machine Translation (NMT), and trained on OPUS data using OPUS-MT-Train. The translation achieves accurate results in machine translation. Next, since articles are translated into English, we process the articles to generate automatic text summaries using mT5 which inherits all the benefits of T5 \cite{raffel2019exploring}. The automatic text summaries currently are English. Finally, we translate automatic text summaries into Arabic by similar settings described in the second step.
% At this stage the summaries are available for the last step, and translated into the source language (Arabic) using the previous similar settings. 
An example of the ground-truth summary and a generated Arabic summary are displayed in Table~\ref{sample_com}.

Both summaries are evaluated by ROUGE \cite{ganesan2018rouge} evaluation metric and will be used for human evaluation (see sec~\ref{human_eval}).  We apply ROUGE-1, ROUGE-2, and ROUGE-L to consider different summary lengths. Moreover, we also show how stemming impacts the accuracy. The results are reported in Table~\ref{rouge}. The results show that the summaries generated by mT5 achieve lower scores before applying the lemmatization process. After we lemmatized the summaries by Farasa, the results improve by a good margin. In both cases, for a model that has not been designed for Arabic language, mT5 shows good scores when scored with LANS summaries see Table~\ref{sample_com}.

\begin{table}[]
\caption{Results of the generated summaries referenced to LANS summaries.}
  \label{rouge}
  \centering
\begin{tabular}{lllllll}
\toprule
    & \multicolumn{3}{l}{Before Lemmatization} & \multicolumn{3}{l}{After Lemmatization} \\
    \midrule
    & R-1         & R-2          & R-L         & R-1         & R-2         & R-L         \\
mT5 & 0.3         & 0.12         & 0.28        & 0.44        & 0.19        & 0.38  \\
\bottomrule
\end{tabular}
\end{table}

\section{Intrinsic Evaluation of LANS}
\label{evaluation}
We apply two methods of evaluation to validate the reliability of the summaries from LANS. The first is an automatic evaluation which examines the summarization techniques in LANS. It uses the following metrics:  \textit{compression ratio}, \textit{fragment density}, and \textit{coverage}. The automatic evaluation has been performed on the whole dataset.
The second evaluation is performed by experts which verifies the quality of LANS by randomly extracting 1000 articles and their respective summaries, which are evaluated by experts. 
% For the former method, the evaluation has been performed on the whole dataset. For the latter, the evaluation set is randomly sampled from LANS for human summary evaluation.

\begin{table}[]
\scriptsize
\begin{center}
        \centering
    \begin{tabular}{llll|llll}
    \hline
        Dataset & \textbf{COV}$\downarrow$ & \textbf{DENS}$\uparrow$ & \textbf{CMP}$\uparrow$ & Dataset & \textbf{COV}$\downarrow$ & \textbf{DENS}$\uparrow$ & \textbf{CMP}$\uparrow$ \\ \hline\hline
        Elkhabar(Algeria)  & 0.34 & 0.87 & 0.77 & Alwatan(Oman) & 0.35 & 0.64 & 0.68 \\ 
        Alwasat(Bahrain) & 0.32 & 0.88 & 0.51 & Alquds(Palestine)  & 0.28 & 0.74 & 0.65  \\ 
        Gate Ahram(Egypt)  & 0.27 & 0.81 & 0.57 & Alquds-UK(Palestine) & 0.39 & 0.90 & 0.79  \\ 
        Youm7(Egypt) & 0.31 & 0.86 & 0.53 & Alwatan(Qatar) & 0.24 & 0.58 & 0.74  \\ 
        Aldustoor(Jordan) & 0.25 & 0.52 & 0.50 & Aljazira(Saudi Arabia)  & 0.23 & 0.46 & 0.57  \\ 
        Annahar(Kuwait) & 0.24 & 0.57 & 0.72 & Alryiadh(Saudi Arabia) & 0.30 & 0.73 & 0.51  \\ 
        Almadapaper(Iraq) & 0.45 & 0.52 & 0.64 & Alsudan Alyoom(Sudan) & 0.36 & 0.31 & 0.49  \\ 
        Alakhbar(Lebanon) & 0.27 & 0.49 & 0.82 & Zamanalwsl(Syria) & 0.26 & 0.62 & 0.59  \\ 
        WAL(Libya)  & 0.32 & 0.30 & 0.55 & Alssabah(Tunisia) & 0.26 & 0.70 & 0.58  \\
        Sahara Media(Mauritania) & 0.32 & 0.88 & 0.68 & Albayan(Emirates) & 0.41 & 0.35 & 0.65 \\
        Hespress(Morocco)  & 0.38 & 1.01 & 0.78 & Almasdar(Yemen) & 0.38 & 0.92 & 0.77 \\\hline
    \end{tabular}
    \caption{Automatic evaluation results of LANS comparing all newspapers to each other. The up arrow $\uparrow$ indicates that higher is better and the opposite for the down arrow $\downarrow$. The results show the diversity among the collected datasets from one source to another. It also shows there is a high level in abstractiveness and conciseness.    }
  \label{eval_table}
 \end{center}
\end{table}

\subsection{Automatic Evaluation}
\label{auto_eval}
To assess LANS, we apply 3 common metrics to quantify the abstractness of LANS's summaries and examine their strategies. Note that summaries can be \textit{extractive} or \textit{abstractive}; extractive summaries derive words from the source text, while abstractive summaries use novel words to describe the source text. The applied metrics used are \textit{compression ratio, fragment density (abstractivity), and coverage} \cite{grusky2018newsroom,bommasani2020intrinsic}. \textbf{Compression Ratio} quantifies the conciseness of summaries, and is defined as the ratio of words between a summary and an article:

\begin{equation}
\label{eq:CMP}
\mathbf{CMP}_{w}(S,A)=1 - {|S|\over{|A|}}
\end{equation}
where $|S|$ is the summary's length and $|A|$ is the article's length in words. \textbf{Coverage} by \cite{grusky2018newsroom} quantifies how much the summary borrows words from the article. Its formula is below: 

\begin{equation}
\label{eq:COV}
\mathbf{COV}(S,A)={1\over{|S|}} \sum_{t \in T(S,A)} |t|
\end{equation}
where $T(S, A)$ is the set of extractive phrases in summary $S$ extracted from article $A$, and	 $t$ is the summary tokens (words) derived from the article. In abstractive summaries, it is preferred not to derive many words from the article.

\textbf{Fragment Density} is proposed by \citep{grusky2018newsroom}, and later introduced as \textbf{Abstractivity} in \citep{bommasani2020intrinsic} with a slight change that generalizes it. This paper uses fragment density. It quantifies how well the summaries can construct a sequence of words that are greedily matched in the article. It is measured as the following:  

\begin{equation}
\label{eq:DENS}
\mathbf{DENS}(S,A)={1\over{|S|}} \sum_{t \in T(S,A)} |t|^{2}
\end{equation}

The results of the automatic evaluation are reported in Table~\ref{eval_table}. The $\downarrow$ arrow for coverage scores (COV) indicates how abstractive the summaries are from each source. The reported low scores signify that the summaries have novel words to describe the articles. The $\uparrow$ arrows for density (DENS) and fragment compression (CMP) mean the higher the better. The highest score for density is in Hespress(Morocco) newspaper summaries, and the lowest is in WAL (a Libyan news agency). For compression, the most concise summaries are reported from Alakhbar (Lebanon), and the least concise ones are reported from Alsudan Alyoom (Sudan). The diversity exists among the Arab countries' style of writing the summaries, and the indication of that is the varying scores in all metrics. %The detailed distributions of \textit{fragment density} and \textit{coverage} across LANS dataset are displayed in the appendix Figure~\ref{fig:fig}

\subsection{Human Evaluation} 
\label{human_eval}
Relying on only automatic evaluation and ROUGE metric may result in some limitations, such as biases in scoring against the systems that depend more on paraphrasing such as abstractive systems\cite{grusky2018newsroom}. As a result, even though meaningful summaries are generated, ROUGE can be subjective and assigns a low score to well-generated summaries\cite{see2017get}. Therefore, we conduct human evaluation.

Human evaluation is costly, but the results from the automatic method described in Sec.~\ref{auto_eval} are yet to be verified by experts. A survey is created for human experts to assess which summaries capture the full \textbf{key information of the articles}, have better \textbf{readability}, and have \textbf{syntactic correctness}. The survey contained the 1000 random samples selected for the experiment in Sec.~\ref{experiment}. Each survey question contains the following data: the full article; Choice 1: LANS summary; Choice 2: mT5-based generated summary; and Choice 3: none-of-the-above (non of the summaries). Choices 1 and 2 were shuffled and anonymized, so human experts can make fairer choices with less biases. For example, if Choice 1 was always LANS's summary, then human experts may form a judgment to always choose Choice 1. Therefore, the choices were shuffled. Besides, the choices were anonymous. It means that human evaluators do not know the origin of each summary. 

The experts who did the survey are highly knowledgeable in Arabic. For a human expert to evaluate the survey; an expert should be an Arabic native speaker, also, an expert should at least have a bachelor's degree majoring in Arabic Language. The experts were asked not only to choose which choice is the fittest for the given criteria, but also to provide their feedback on the choices.  
Human evaluation results show that 954 of LANS extracted summaries have more accurate semantic representation, and correct syntactic forms. 
The semantic representation means that the summary captures salient and key information of the article and has better readability. The results, also, show that 2 of the choices are "none", which means neither summaries meet the required criteria. While the ROUGE scores are low between the automatic generated summaries and LANS summaries, the human evaluation results verify the correctness of LANS summaries with an accuracy of 95.4\%. 

\section{Conclusion} 
\label{conclusion}
%LANS overcomes the limitations of the existing ATS datasets in terms of diversity and size. 
This work presents LANS, a large-scale and diverse text summarization dataset of more than 8 million new articles (32GB) paired with their summaries written by journalists. The summaries are collected from the metadata of 22 scraped popular Arab newspapers' websites from the period between 1999 to 2019. For each of those resources, LANS considered a wide range of topics. The work applied two evaluation methods (automatic and human) to verify the superiority of the extracted summaries in LANS. The dataset is available on this link~\footnote{Request data from first author}. LANS offers this dataset for researchers to advance the field of ATS, and takes advantage of the data to train and evaluate the results of new models on this dataset. For future work, we plan to benchmark LANS for text classification since the articles' labels are available.

\bibliographystyle{unsrtnat}
\bibliography{template}  %%% Uncomment this line and comment out the ``thebibliography'' section below to use the external .bib file (using bibtex) .
% \bibliography{references}





% \end{thebibliography}

\end{document}